\documentclass[conference]{IEEEtran}
\IEEEoverridecommandlockouts
\usepackage{cite}
\usepackage{amsmath,amssymb,amsfonts}
\usepackage{algorithmic}
\usepackage[ruled,linesnumbered]{algorithm2e}
\usepackage{graphicx}
\usepackage{textcomp}
\usepackage{xcolor}
\usepackage{listings}
\usepackage{xcolor}
\usepackage{booktabs}
\usepackage{multirow}  
\usepackage{graphicx}
\usepackage{array}   
\usepackage{hyperref}
\usepackage{amsmath}
\DeclareMathOperator*{\argmin}{arg\,min}
\lstdefinelanguage{Verilog}{
  keywords={module, input, output, wire, reg, always, begin, end, if, else},
  keywordstyle=\color{blue}\bfseries,
  comment=[l]{//},
  commentstyle=\color{gray}\ttfamily,
  sensitive=true
}

\def\BibTeX{{\rm B\kern-.05em{\sc i\kern-.025em b}\kern-.08em
    T\kern-.1667em\lower.7ex\hbox{E}\kern-.125emX}}
\begin{document}

\title{Agent Factories for High Level Synthesis:\\How Far Can General-Purpose Coding Agents Go in Hardware Optimization? \\
}

\author{\IEEEauthorblockN{Abhishek Bhandwaldar, Mihir Choudhury, Ruchir Puri, Akash Srivastava} \textit{IBM Corporation} }

\maketitle
\begin{abstract}
We present an empirical study of how far general-purpose coding agents---with no hardware-specific training---can go in optimizing hardware designs from high-level algorithmic specifications. Our method, an \emph{agent factory}, is a two-stage pipeline that builds and coordinates multiple autonomous optimization agents. In Stage~1, the factory decomposes a design into sub-kernels, independently optimizes each through pragma and code-level transformations, and solves an Integer Linear Program to assemble globally promising configurations under area constraints. In Stage~2, it spawns $N$ expert agents over the top ILP solutions, each exploring cross-function optimizations---pragma recombination, loop fusion, memory restructuring---that sub-kernel decomposition cannot reach.

We evaluate the pipeline on twelve kernels from HLS-Eval and Rodinia-HLS using Claude Code (Opus~4.5/4.6) with AMD Vitis~HLS. Scaling from 1 to 10~agents yields a mean $7.07\times$ speedup over baseline, with gains concentrated on harder problems: \textit{leukocyte} and \textit{kmeans} reaches ${\sim}10\times$. Across benchmarks, agents consistently recover known hardware optimization patterns without domain-specific training, and winning designs frequently originate from non-top-ranked ILP variants, confirming that global optimization uncovers improvements invisible to sub-kernel search. These findings establish agent scaling as a practical axis for HLS optimization. We provide an anonymized implementation \href{https://anonymous.4open.science/r/anonymous-sub-E347/}{here}.
\end{abstract}


\section{Introduction}
\label{sec:intro}

High-Level Synthesis (HLS) strives to raise the abstraction of hardware design from
RTL to C/C++, with the goal of achieving high performance. However, with current
state-of-the-art HLS tools, significant expert-driven
pragma insertion and code restructuring remains necessary to achieve desirable
results~\cite{autodse,opentuner,comba}.
Selecting the right combination of directives such as \texttt{PIPELINE},
\texttt{UNROLL}, \texttt{ARRAY\_PARTITION}, and others requires deep
hardware knowledge, careful latency-area tradeoff reasoning, and extensive
iteration with the synthesis toolchain~\cite{cong2022hls}.
As an example, Cong et al.\ report that more than 40\%
of lines of high-level code in a real-world genomics kernel were attributable
to hardware-specific optimizations and pragmas alone~\cite{cong2022hls},
illustrating the manual effort that persists even with modern HLS tools.

\begin{figure}[t]
\centering
\includegraphics[width=\columnwidth]{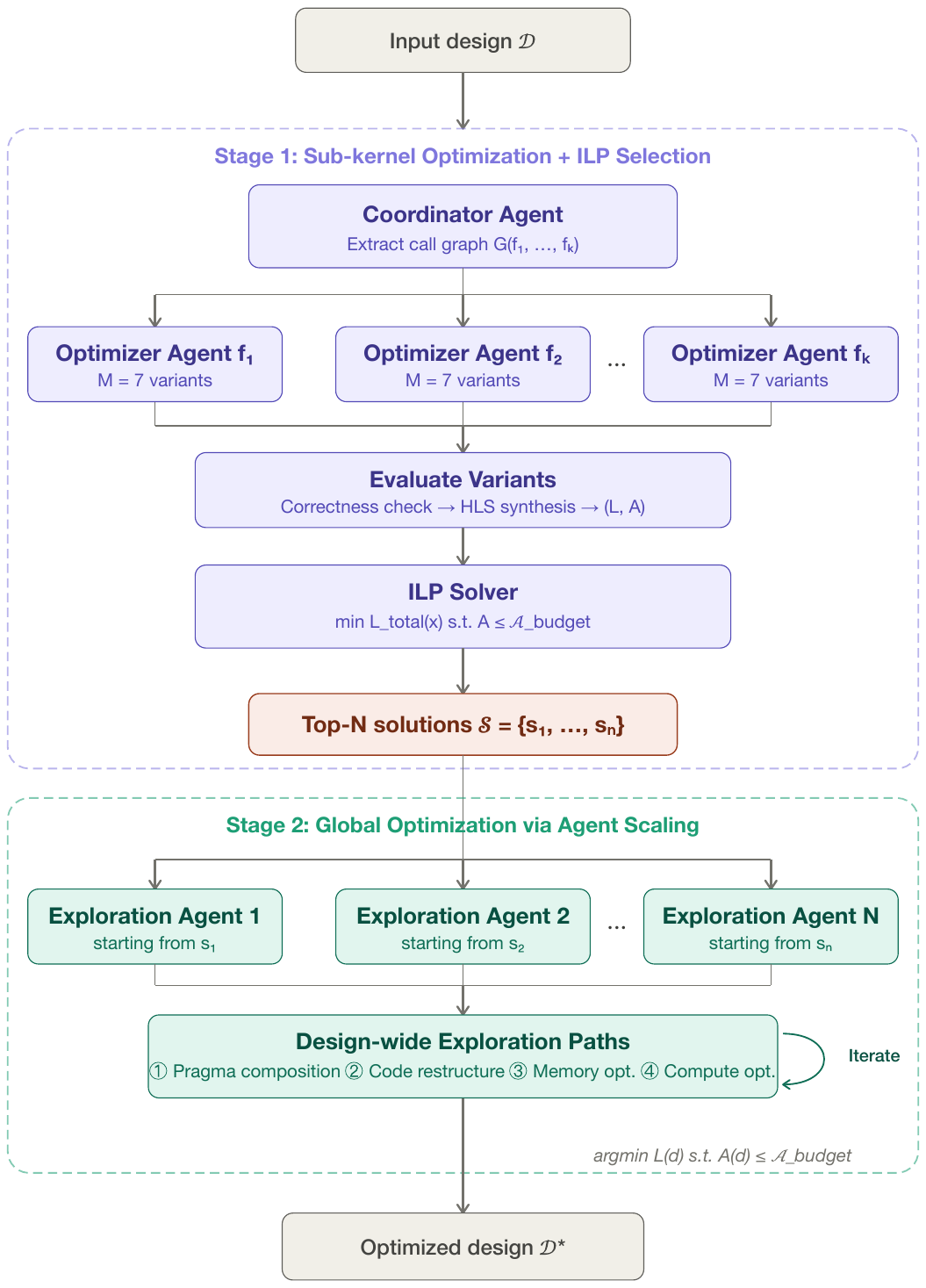}
\caption{Two-stage agent-based pipeline for HLS design space exploration. Given an input design $\mathcal{D}$, a coordinator agent extracts the function call graph $G$ and spawns one optimizer agent per sub-function $f_1, \dots, f_K$. Variants are evaluated for correctness and synthesized to obtain (latency, area) pairs. An ILP solver then selects the top-$N$ combinations $\mathcal{S} = \{s_1, \dots, s_N\}$ that minimize total latency subject to the area budget. In Stage~2, $N$ exploration agents each start from a candidate solution and iteratively apply design-wide optimization paths, to produce the final optimized design $\mathcal{D}^*$.}
\label{fig:placeholder}
\vspace{-10pt}
\end{figure}

Existing automation approaches treat {\textbf{high-level code to RTL generation}} as a black-box optimization
problem over a predefined parameter space.
Bayesian optimization methods~\cite{autohls,hgbodse,Kuang2024CompassAC} build
surrogate models over pragma configurations;
Integer Linear Programming (ILP) and non-linear programming formulations~\cite{ilp3,pouget2025pragma}
compose optimized configurations from pre-enumerated candidates.
These approaches navigate large configuration spaces efficiently, but
cannot restructure code, rewrite algorithms, or discover optimizations
outside the predefined parameter space.
Recent LLM-based methods~\cite{lift2025,hlspilot,reasoning_hls}
are a promising step, but most still operate within structured pragma
selection rather than open-ended program transformation.

This work explores \emph{how far general-purpose coding
agents---with no HLS-specific training---can go in optimizing hardware designs,
given only source code, synthesis tool access, and freedom to modify both
code and pragmas.}
The motivation is straightforward: recent agentic coding
systems~\cite{yang2024sweagent,hong2024metagptmetaprogrammingmultiagent,qian2024chatdevcommunicativeagentssoftware,wong2026confuciuscodeagentscalable}
have shown that LLM-based agents can iteratively refine code using tool
feedback across a range of software engineering tasks, and similar
agent-driven optimization efforts are emerging in adjacent
domains~\cite{karpathy2026autoresearch}.
The key question is whether this capability transfers to the hardware
domain, where optimization requires reasoning about latency, area, memory
bandwidth, and non-obvious pragma interactions.
Unlike parameter-sweep methods, agents search over \emph{programs}: in
principle, they can restructure loops, replace computations with closed-form
expressions, or reorganize memory access. In practice, as our results show,
agents most often discover effective pragma strategies, but occasionally
apply code-level transformations that go beyond what predefined parameter
spaces can enumerate.

Our framework uses an \emph{agent factory} that generates multiple autonomous
optimization agents exploring different trajectories through the design space.
The framework operates in two stages: (1)~sub-kernels are independently
optimized and combined via ILP under area constraints;
(2)~$N$ global agents explore full-design transformations over the
top-ranked ILP solutions.
By varying $N$, we study whether agent scaling, allocating more
inference-time compute, improves solution quality.
We evaluate the pipeline on 12 kernels: 6 from prior HLS
benchmarks~\cite{reasoning_hls} and 6 from
Rodinia-HLS~\cite{10.1145/3174243.3174970}, using Claude Code
(Opus~4.5/4.6) and AMD Vitis~HLS.

We frame this as an \textbf{empirical study}: our benchmark set is small,
several kernels are well-studied, and results are specific to a single model
family and toolchain. The contribution is a systematic characterization of
what general-purpose agents can and cannot do in hardware optimization, with
the goal of establishing a baseline that the community can build upon. Our
findings:

\textbf{1. Agent scaling improves exploration.}
Increasing the number of agents from 1 to 10 yields a mean $8.27\times$ speedup
over baseline, with the largest gains on complex workloads:
\textit{streamcluster} exceeds $20\times$, and \textit{kmeans} reaches
${\sim}10\times$. Simpler or pipeline-dominated kernels show limited or
saturating improvements.

\textbf{2. Agents go beyond isolated pragma selection.}
The best final design does not always originate from the top-ranked ILP variant,
indicating that global optimization across function boundaries can uncover
improvements not reachable by sub-kernel search alone. On \textit{lavamd},
agents achieve ${\sim}8\times$ speedup at ${\sim}40$--$60$K area, improving
area--latency trade-offs compared to reference implementations.

\textbf{3. Agents recover known hardware optimization patterns without training.}
Across kernels, agents consistently apply \texttt{ARRAY\_PARTITION} to
resolve memory bottlenecks and learn that \texttt{PIPELINE} is ineffective
unless loop-carried dependencies are first addressed, patterns that align
with established HLS expertise.

\textbf{4. Limitations are significant and inform future work.}
This is a preliminary study with inherent scope constraints. The benchmark
set of twelve kernels does not capture the full complexity of real-world
HLS workloads. All experiments use a single model family (Claude
Opus~4.5/4.6), a single synthesis tool (Vitis~HLS), and a single target
architecture (FPGA). Baselines, while systematic, are bounded exhaustive
searches over restricted directive sets rather than comparisons against
state-of-the-art DSE frameworks such as AutoDSE. Gains are also uneven:
simpler kernels saturate early, and under tight area budgets additional
agents can increase area without proportional latency improvement. We
view this study as a starting point intended to bootstrap a line of
investigation that can be extended, through broader benchmarks, stronger
baselines, additional models, and diverse target architectures, via
community contribution.

\section{Related Work}
\label{sec:related}

\paragraph{Automated Design Space Exploration for HLS}
Optimizing HLS designs requires tuning directives such as pipelining, unrolling, and memory partitioning, motivating extensive work on automated DSE. Early approaches relied on heuristic and analytical strategies~\cite{cong2011hls,cong2012dse}, followed by iterative synthesis-driven frameworks~\cite{autodse}, lattice-based traversal~\cite{latticehls}, multi-level parallelism modeling~\cite{MPseeker}, and multi-strategy autotuning~\cite{opentuner,armbanditdse,sherlockdse}. Learning-based methods accelerate DSE through surrogate models~\cite{autodse,comba,chang2023dr,Bai_2022}, GNN-based program representations~\cite{10.1145/3489517.3530409,harp,Wu_2021}, and cross-modality learning~\cite{qin2024crossmodalityprogramrepresentationlearning}. These approaches largely treat optimization as search over a predefined parameter space, limiting their ability to perform open-ended program transformations or capture global interactions across sub-kernels.

\paragraph{LLM-based Optimization for HLS}
Recent work applies LLMs to HLS optimization along three directions. First, directive generation systems such as HLSPilot, LIFT, LLM-DSE, and iDSE use LLMs to propose pragma configurations within synthesis-in-the-loop feedback~\cite{hlspilot,lift2025,wang2025llmdsesearchingacceleratorparameters,li2025idsenavigatingdesignspace}, but remain within predefined parameter spaces. Second, source-to-source transformation approaches iteratively convert C/C++ into synthesizable HLS code, focusing primarily on correctness and synthesizability~\cite{c2hlsc,xu2024automatedccprogramrepair}. Third, agentic pipelines integrate profiling, transformation, and DSE~\cite{hlspilot,oztas2024agentichlsagenticreasoningbased}, but typically follow a single optimization trajectory or fixed coordination strategy.

\paragraph{Agentic LLM Systems and Scaling}
LLM-based agents enable multi-step reasoning, tool use, and iterative code refinement across software engineering tasks~\cite{yang2024sweagent,hong2024metagptmetaprogrammingmultiagent,qian2024chatdevcommunicativeagentssoftware,wong2026confuciuscodeagentscalable}. Our work builds on this paradigm by treating the number of agents as inference-time compute, analogous to test-time scaling~\cite{2025-its}. While prior agentic systems typically rely on a small number of coordinated agents, we study \emph{agent scaling} as a first-class design dimension for HLS optimization.

\section{Method}

Given an input program written in a high-level language (e.g., C/C++) and a target implementation such as an FPGA platform or ASIC with resource limits, the goal of HLS design space exploration is to find an implementation that minimizes execution latency (in clock cycles) while satisfying area constraints, flip-flops, RAMs, and DSP blocks. The design space consists of all feasible combinations of HLS pragma assignments (e.g., \texttt{PIPELINE}, \texttt{UNROLL}, \texttt{ARRAY\_PARTITION}) and code-level transformations (e.g., loop restructuring, memory reorganization) applied to the input program, where each configuration yields a distinct hardware implementation with different latency and resource utilization.

This space is difficult to explore for three reasons:
\begin{enumerate}
    \item First, the number of feasible configurations grows combinatorially with the number of loops, arrays, and functions, and each configuration requires an HLS synthesis run that may take minutes, making exhaustive search impractical.
    \item Second, optimization decisions interact globally: aggressively optimizing one function may exhaust the area budget and prevent improvements elsewhere, so exploration must balance decisions across the full design.
    \item Third, the effect of pragmas is non-linear and sometimes counterintuitive -- for example, in the Needleman--Wunsch kernel, fully unrolling the \texttt{reverse\_string} loop increased latency from 26 to 71 cycles due to memory port contention, illustrating that more aggressive optimization does not always improve performance.
\end{enumerate}

These challenges motivate a two-stage pipeline that decomposes the problem into independent sub-kernel optimization followed by global, design-wide exploration.

\subsection{Notation}
\label{sec:notation}

Let the input design $\mathcal{D}$ consist of a top-level function that invokes a set of
sub-functions $\mathcal{F}=\{f_1,f_2,\dots,f_K\}$.
We extract a \emph{function call graph} $G=(\mathcal{F},\mathcal{E})$, where a directed edge
$(f_i,f_j)\in\mathcal{E}$ indicates that $f_i$ invokes~$f_j$.
From $G$ we compute the critical path and classify inter-function dependencies as
sequential or parallel.

For each sub-function $f_k$, we define a \emph{variant set}
$\mathcal{V}_k=\{v_k^{0},v_k^{1},\dots,v_k^{M}\}$, where each variant $v_k^{m}$ is a
distinct optimization configuration (pragma assignment and/or code transformation).
Each variant is characterized by its latency $L_k^{m}$ and area $A_k^{m}$, obtained
through HLS synthesis after passing a functional correctness check.
A binary selection variable $x_k^{m}\in\{0,1\}$ indicates whether variant~$m$ of
sub-function~$k$ is selected.

Let $\mathcal{A}_{\mathrm{budget}}$ denote the global area constraint imposed by the target
FPGA platform, and let $L_{\mathrm{baseline}}$ denote the latency of the unoptimized design.

\subsection{Stage~1: Sub-Kernel Optimization and ILP Selection}
\label{sec:stage1}

Stage~1 proceeds in three phases: variant generation, evaluation, and ILP-based selection.

\subsubsection{Phase~1a --- Variant Generation}
A coordinator agent analyzes the call graph $G$ and spawns one \emph{optimizer agent}
per sub-function $f_k\in\mathcal{F}$.
Each optimizer explores $M=7$ variants following a structured strategy
(Table~\ref{tab:variants}).

\begin{table}[t]
  \centering
  \caption{Variant exploration strategy per sub-function.}
  \label{tab:variants}
  \renewcommand{\arraystretch}{1.3} 
  \begin{tabular}{cl}
    \toprule
    Variant & Strategy \\
    \midrule
    $v_k^{0}$ & Baseline: synthesize unmodified code \\
    $v_k^{1}$ & Conservative: minimal pragmas, low area \\
    $v_k^{2,3}$ & Pipeline: \texttt{PIPELINE II}\,$\in\{1,2,4\}$ \\
    $v_k^{4,5}$ & Aggressive: pipeline + partial/full \texttt{UNROLL} \\
    $v_k^{6}$ & Alternate: \texttt{ARRAY\_PARTITION}, \texttt{INLINE}, \\
               & \quad closed-form rewrites \\
    \bottomrule
  \end{tabular}
\end{table}

\subsubsection{Phase~1b --- Evaluation}
Each variant $v_k^{m}$ is evaluated through a pipeline of
(i)~functional correctness testing against the original design,
(ii)~HLS synthesis on the target platform, and
(iii)~extraction of latency $L_k^{m}$ and area $A_k^{m}$.
Variants that fail correctness are discarded.

\subsubsection{Phase~1c --- ILP Formulation}
\begin{figure*}[t]
\centering
\includegraphics[width=.8\textwidth]{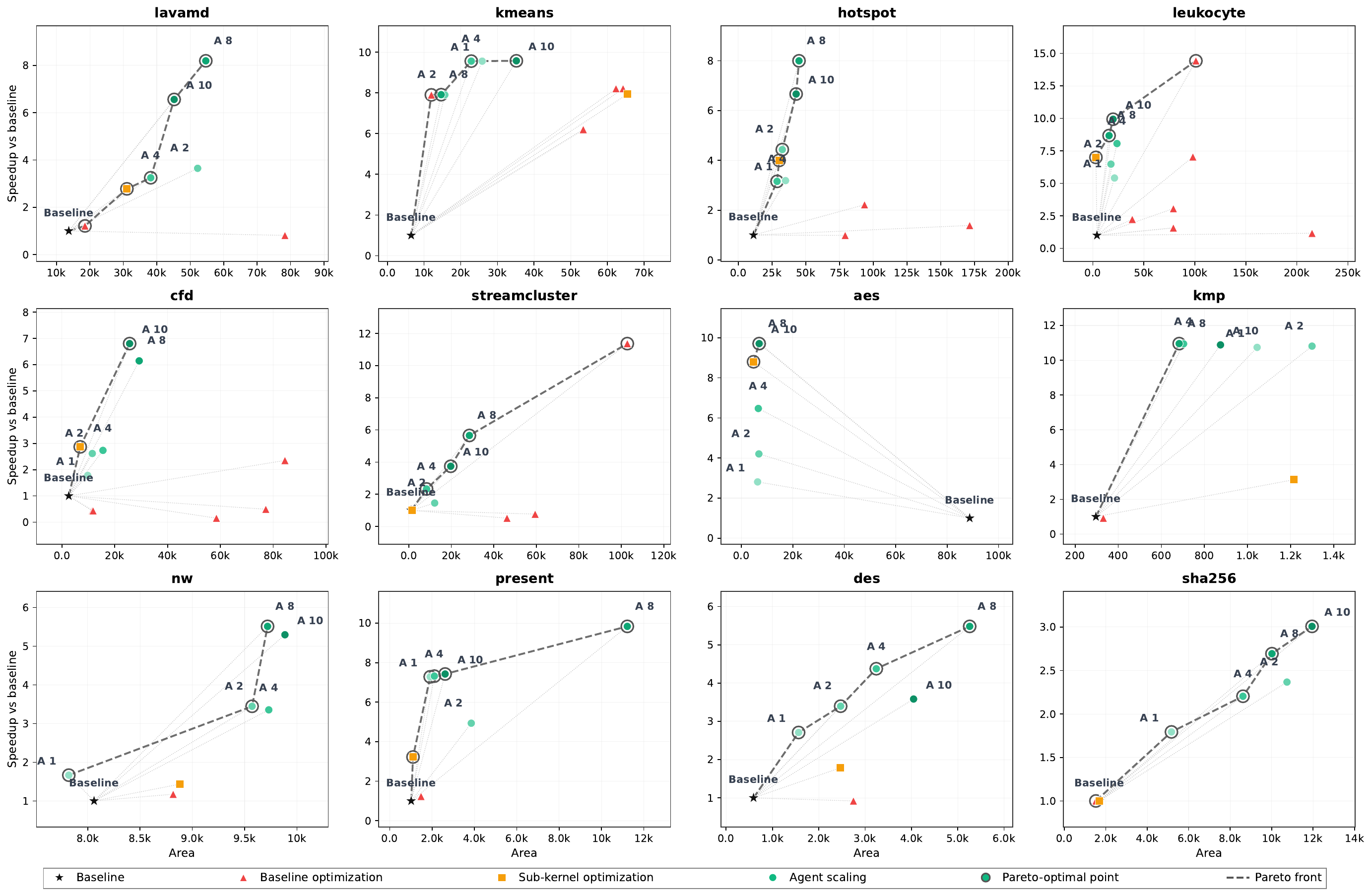}
\caption{Pareto front results for all twelve benchmarks under agent scaling
  ($N \in \{1,2,4,8,10\}$). Each subplot shows speedup over baseline
  (y-axis) versus area (x-axis). Increasing the number of agents extends
  the Pareto front toward lower latency and more favorable area--latency
  trade-offs across most benchmarks, with the strongest gains on harder
  problems such as streamcluster, leukocyte, and NW.}
\label{fig:pareto}
\end{figure*}
Once all optimizer agents complete, the coordinator aggregates the surviving
variants for each sub-kernel. At this stage, each function $f_k$ has multiple
candidate implementations with latency $L_k^{m}$ and area $A_k^{m}$ obtained
from HLS synthesis.

Selecting the best variant for each function independently is insufficient,
because the latency of the overall design depends on how these functions
compose in the program execution structure. In HLS designs, functions may
execute sequentially, overlap through pipelining, or appear within loop
iterations. As a result, the global latency cannot be expressed as a simple
sum of individual function latencies.

To determine how candidate variants interact in the full design, the
coordinator analyzes the program call graph $G$ and derives a latency
composition model that reflects the execution structure of the program.
For sub-functions on a sequential path, latencies accumulate; for sub-functions
that execute in parallel regions, the dominant latency determines the stage
duration. The resulting latency model is expressed as 
\begin{equation}
     L_{\mathrm{total}}(\mathbf{x})
  \;=\;
  h\!\bigl(\{L_k^{m}\cdot x_k^{m}\}_{k,m}\bigr),
  \label{eq:latency-composition} 
\end{equation}

where $h(\cdot)$ encodes the latency composition derived from the call graph
(e.g., sums along sequential chains and maxima over parallel branches, with
appropriate loop multipliers). The binary decision variable $x_k^{m}$ indicates
whether variant $m$ of sub-function $f_k$ is selected.

Using this derived latency model, the coordinator constructs an Integer Linear
Programming (ILP) formulation that selects one variant per sub-kernel while
respecting the global area constraint:

\begin{align}
  \min_{\mathbf{x}}\; & L_{\mathrm{total}}(\mathbf{x}) \nonumber \\
  \text{s.t.}\;\;
  & \textstyle\sum_{m} x_k^{m} = 1 \;\;\forall\, f_k \in \mathcal{F}, \nonumber \\
  & \textstyle\sum_{k,m} A_k^{m}\,x_k^{m} \leq \mathcal{A}_{\mathrm{budget}}, \quad
  x_k^{m} \in \{0,1\}.
  \label{eq:ilp}
\end{align}

Because multiple variant combinations can satisfy the area constraint while
achieving similar latency, the solver is used to enumerate the top-$N$
feasible solutions ranked by $L_{\mathrm{total}}$, denoted
$\mathcal{S}=\{s_1,s_2,\dots,s_N\}$. Each solution corresponds to a distinct
global configuration of sub-kernel implementations and serves as a starting
point for the next phase of full-design optimization.

\subsection{Stage~2: Global Optimization via Agent Scaling}
\label{sec:stage2}

Stage~2 takes the candidate set $\mathcal{S}$ from the ILP solver and attempts to
improve each candidate through holistic, design-wide optimization.

A coordinator agent spawns $N$ \emph{exploration agents}, one per candidate
$s_i\in\mathcal{S}$.
Each agent receives the full design instantiated with the sub-kernel variants
specified by $s_i$ and explores optimization paths that operate across function
boundaries.  These paths are categorized as:

\textbf{1. Pragma composition:} new combinations of HLS pragmas applied jointly across multiple functions, not explored during Stage~1.

 \textbf{2. Code restructuring:} loop reordering, loop fusion, or function
        inlining at the global level.
        
 \textbf{3. Memory optimization:} cross-function array partitioning and memory
        access restructuring.
        
\textbf{4. Compute optimization:} algebraic simplifications or closed-form
        transformations spanning multiple sub-kernels.

Each agent iteratively generates modified designs, verifies correctness,
synthesizes via HLS, and records (latency, area) pairs.
Let $\mathcal{R}_i$ denote the set of all valid designs explored by agent~$i$.
The final output of the pipeline is:
\begin{equation}
  \mathcal{D}^{*}
  \;=\;
  \argmin_{d\,\in\,\bigcup_{i=1}^{N}\mathcal{R}_i}\;
  L(d)
  \quad\text{s.t.}\quad
  A(d)\leq\mathcal{A}_{\mathrm{budget}}.
  \label{eq:final-selection}
\end{equation}

\subsection{Algorithm Summary}

\label{sec:algorithm}

Algorithm~\ref{alg:pipeline} in the appendix summarizes the complete two-stage pipeline.
By parameterizing Stage~2 with $N$, we study how increasing the number of
parallel exploration agents affects solution quality.
Larger $N$ expands the explored region of the design space, increasing the
probability of discovering lower-latency implementations.
We evaluate $N\in\{1,2,4,8,10\}$ in Section~\ref{sec:results}.

\section{Results}
\label{sec:results}

\subsection{Experimental Setup}

We evaluate the two-stage pipeline on twelve HLS kernels: six HLS-Eval benchmarks from prior work~\cite{reasoning_hls} (AES, DES, KMP, NW, PRESENT, SHA256) and six
from Rodinia-HLS~\cite{10.1145/3174243.3174970} (lavamd, kmeans, hotspot,
leukocyte (\texttt{lc\_dilate}), cfd (\texttt{cfd\_step\_factor}), streamcluster).
Results are averaged over 5 runs; when HLS reports latency as a range, we use
the midpoint. All experiments use Claude Code (Opus~4.5/4.6) with AMD Vitis HLS.

\paragraph{Baselines}
For HLS-Eval kernels, we use a bounded exhaustive baseline that enumerates per-loop pragma configurations for each sub-kernel. Each loop selects one of five options: no directive, \texttt{PIPELINE} (II $\in {1,2}$), or \texttt{UNROLL} (factor $\in {2,4}$), capturing common loop optimizations with a manageable branching factor. The search space grows as $5^n$ for $n$ loops, so we cap variants per function to keep synthesis tractable. Each variant is synthesized to obtain latency and area, and an ILP selects one variant per sub-kernel to minimize latency under a global area constraint. We design the baseline to operate at the pragma level, isolating directive search and enabling a controlled evaluation of the additional benefits from higher-level transformations (e.g., code restructuring, memory layout, array partitioning) introduced by the LLM agent. This provides a strong pragma-only reference, though not globally optimal due to the bounded search.

For Rodinia-HLS kernels, we compare against reference optimized implementations
(e.g., tiling, pipelining, double-buffering) and report the best feasible design.

All designs must satisfy area and timing constraints. We use a fixed clock of
10\,ns (35\,ns for \textit{streamcluster}); infeasible baseline configurations
are adjusted by reducing pragma factors, and designs violating area or timing
are discarded.

\subsection{Results on FPGA using Vitis HLS}
\begin{table}[t]
\centering
\small
\caption{Mean speedup over baseline as agents scale}
\label{tab:expert_scaling}
\setlength{\tabcolsep}{6pt}
\begin{tabular}{lcccc}
\toprule
 & \textbf{1$\to$2} & \textbf{2$\to$4} & \textbf{4$\to$8} & \textbf{8$\to$10} \\
\midrule
\textbf{Speedup} &
\shortstack{4.31$\times$\\(+5.9\%)} &
\shortstack{4.92$\times$\\(+14.2\%)} &
\shortstack{7.07$\times$\\(+43.7\%)} &
\shortstack{6.53$\times$\\(-7.6\%)} \\
\bottomrule
\end{tabular}
\end{table}

\subsubsection{Sub-Kernel Optimization}
Across the twelve benchmarks, Stage~1 improves latency over the baseline while
remaining within the area budget; SHA256 is the exception, where performance
stays comparable to baseline (Fig.~\ref{fig:pareto}).

Two recurring patterns emerge without any HLS-specific training.
First, agents consistently identify \texttt{ARRAY\_PARTITION} as the
highest-impact directive, yielding the largest gains in AES, DES, and PRESENT by resolving memory bottlenecks.
Second, agents learn that \texttt{PIPELINE} applied in isolation is often
ineffective---or even harmful---unless memory bandwidth and loop-carried
dependencies are addressed first.

\subsubsection{ILP Variant Selection}
Following sub-kernel optimization, a coordinator agent composes the surviving
variants into a full design. It analyzes the function call graph, derives a
latency composition model that accounts for sequential and parallel execution
paths, and formulates the ILP objective (Eq.~\ref{eq:ilp}) to minimize global
latency under an area constraint.

Table~\ref{tab:ilp_formulation} validates the agent's inferred latency
structure on two synthetic benchmarks (SYN5, SYN6) from~\cite{reasoning_hls}
and two real kernels (NW, AES). The agent correctly captures parallel
compositions---minimizing the maximum latency for concurrent modules
(SYN5)---and mixed parallel-sequential structures (SYN6).

Because this formulation estimates global latency from isolated sub-kernel
synthesis, it cannot capture cross-function effects such as inter-kernel
memory reuse or global pipeline scheduling. Exhaustive enumeration would
resolve this gap but is computationally prohibitive. The ILP stage therefore
serves as an efficient filter that narrows the search space, producing a
ranked set of $N$ starting points for Stage~2.

\begin{table}[t]
\centering
\caption{The agent analyzes data flow graphs and automatically formulates ILP constraints to minimize latency under an area budget. Dataflow graph borrowed from \cite{reasoning_hls}.}
\label{tab:ilp_formulation}
\setlength{\tabcolsep}{3pt}
\footnotesize

\begin{tabular}{m{3.7cm} m{4.4cm}}
\hline
\textbf{DFG} & \textbf{Agent-formulated ILP \& constraints} \\
\hline

\includegraphics[width=3.5cm]{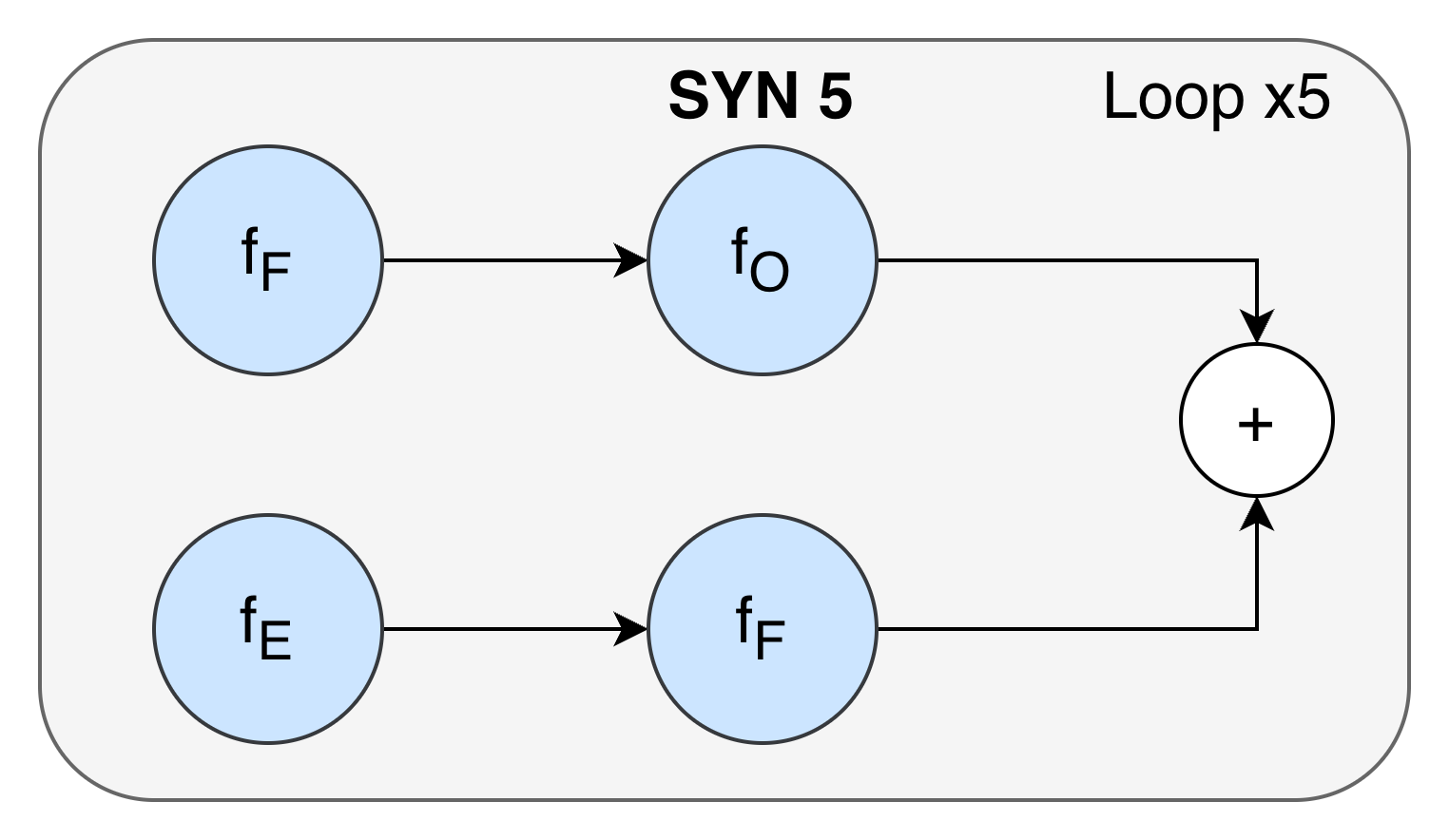}
&
\begin{minipage}[c]{\linewidth}
\vspace{1mm}
$\min L_{iter}\cdot Loop\_count$\\
s.t. $L_{iter}\ge L_F+L_O,\; L_{iter}\ge 2L_F,\; L_{iter}\ge L_E+L_F$\\
$\sum_v x_v=1,\; \sum_v A_vx_v\le A_{budget},\; x_v\in\{0,1\}$\\
\emph{One variant per function; $A_v$ is variant area.}
\vspace{1mm}
\end{minipage}
\\
\hline

\includegraphics[width=3.5cm]{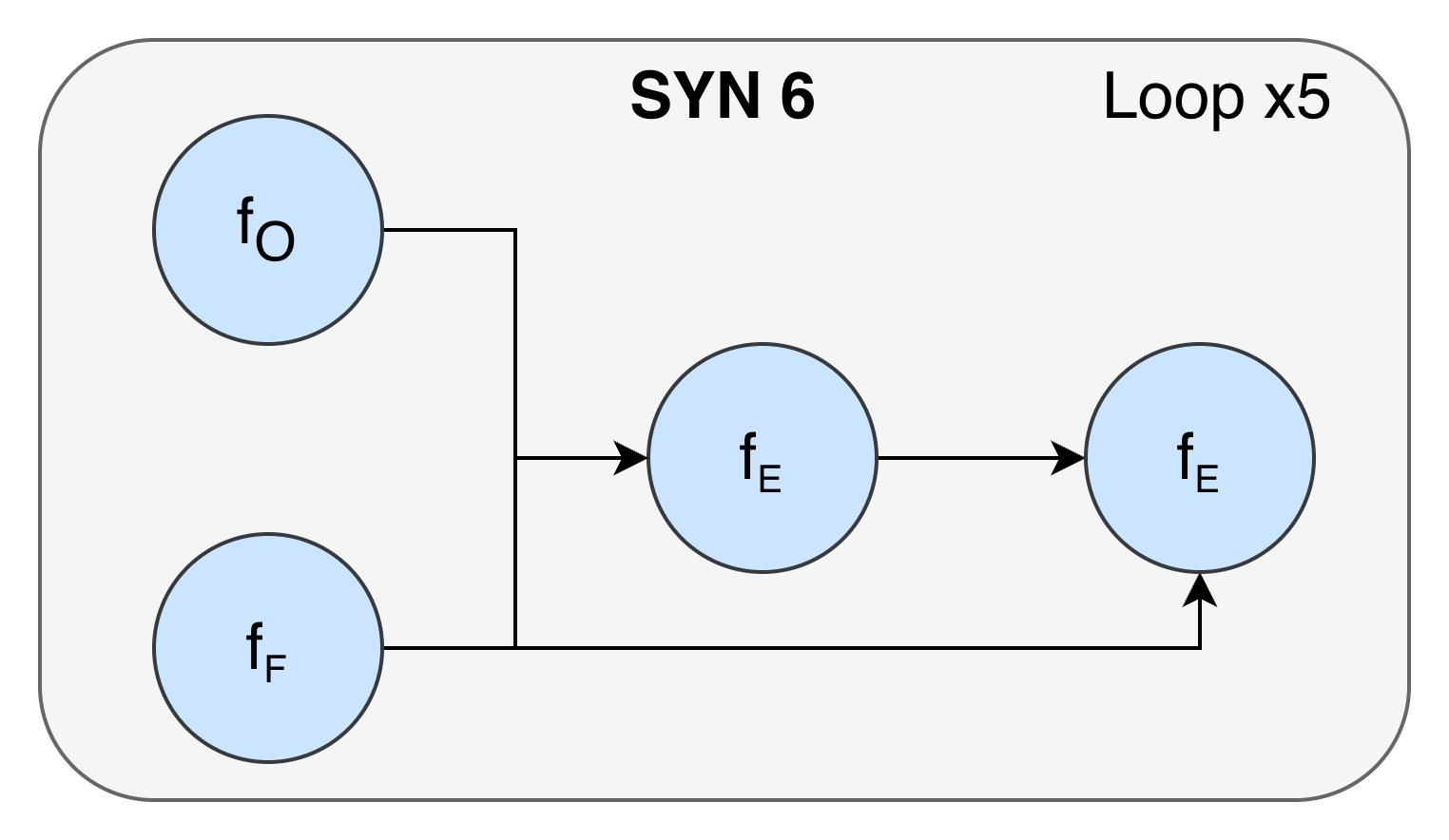}
&
\begin{minipage}[c]{\linewidth}
\vspace{1mm}
$\min 5(M_{FO}+2L_E)$\\
s.t. $M_{FO}\ge L_F,\; M_{FO}\ge L_O$\\
$\sum_v x_v=1,\; \sum_v A_vx_v\le A_{budget},\; x_v\in\{0,1\}$\\
\emph{$L_F,L_O$ denote latencies of F and O.}
\vspace{1mm}
\end{minipage}
\\
\hline

\includegraphics[width=3.5cm]{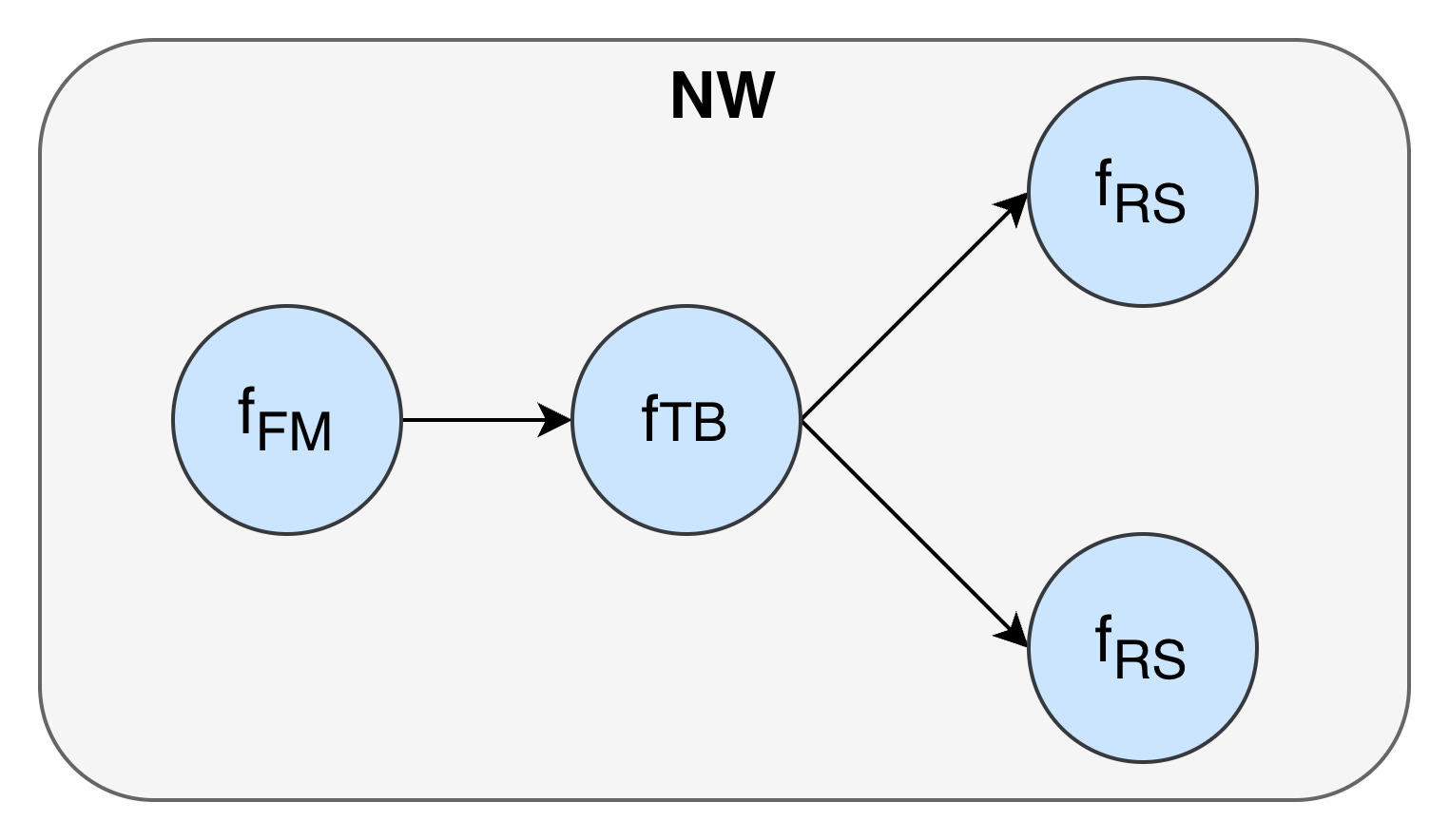}
&
\begin{minipage}[c]{\linewidth}
\vspace{1mm}
$\min (L_{FM}+L_{TB}+L_{RS})$\\
$\sum_v x_v=1,\; \sum_v A_vx_v\le A_{budget},\; x_v\in\{0,1\}$\\
\emph{$L_{FM},L_{TB},L_{RS}$ denote module latencies.}
\vspace{1mm}
\end{minipage}
\\
\hline

\includegraphics[width=3.5cm]{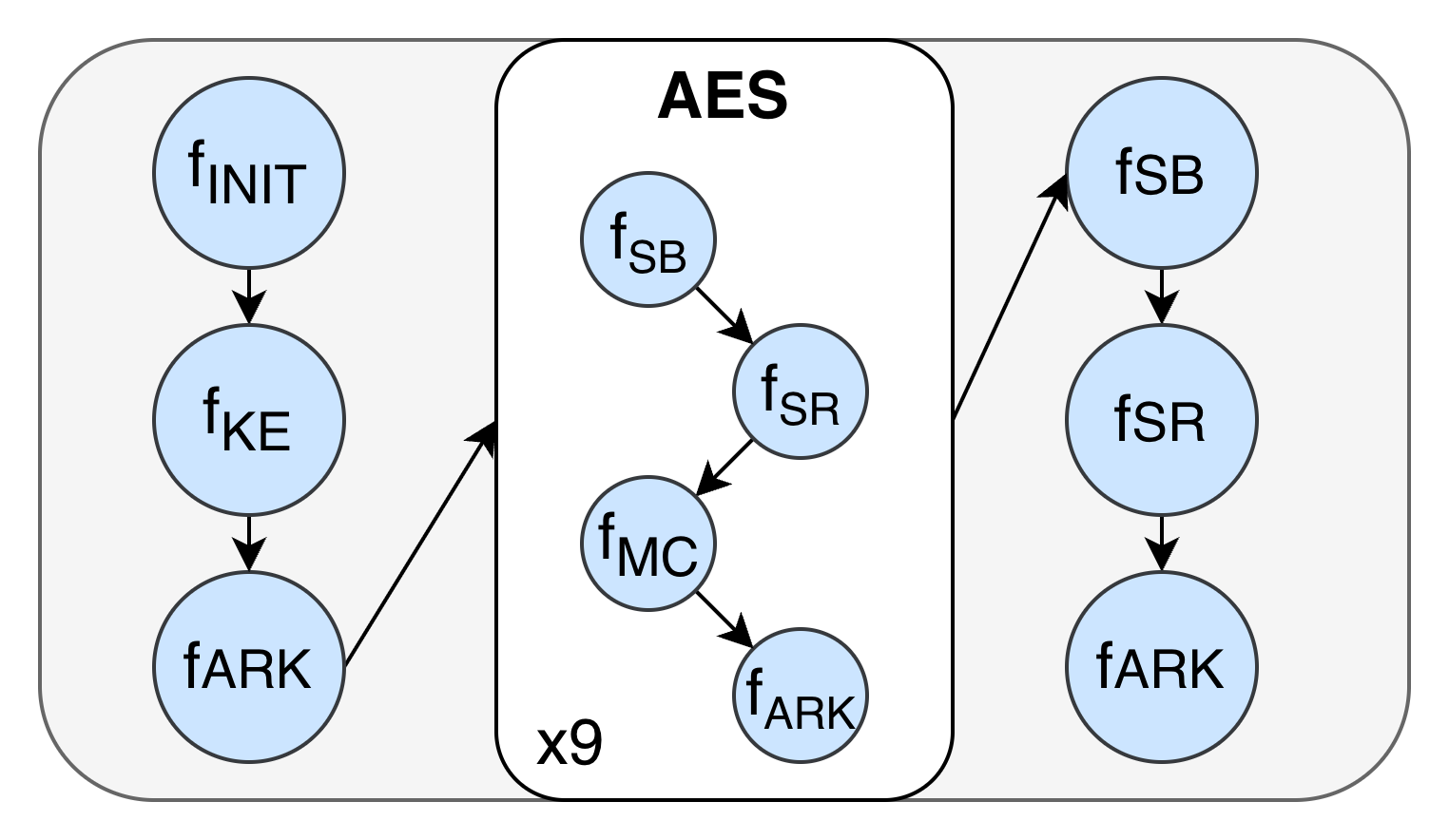}
&
\begin{minipage}[c]{\linewidth}
\vspace{1mm}
$\min 11L_{ARK}+10L_{SB}+10L_{SR}+9L_{MC}+L_{KE}+L_{INIT}$\\
$\sum_v x_v=1,\; \sum_v A_vx_v\le A_{budget},\; x_v\in\{0,1\}$\\
\emph{$L_*$ denotes module latency.}
\vspace{1mm}
\end{minipage}
\\
\hline

\end{tabular}
\end{table}

\subsubsection{Global Optimization}
Figure~\ref{fig:pareto} shows Pareto fronts across all twelve benchmarks, and
Table~\ref{tab:expert_scaling} reports the mean best speedup over 5~runs as
agents scale from $N{=}1$ to $N{=}10$. We organize the findings below by the
character of the scaling behavior.

\paragraph{Strong scaling on complex workloads}
The largest gains appear on benchmarks with rich optimization landscapes,
though improvements are more moderate than initially observed.
\textit{Streamcluster} shows consistent benefits from scaling, with speedup
reaching ${\sim}5$--$6\times$ within the evaluated range of up to 10 agents,
indicating continued gains without clear evidence of saturation.
\textit{Cfd}, \textit{hotspot}, and \textit{kmeans} follow a similar
pattern, reaching $7$--$10\times$ speedup with progressively better
area--latency trade-offs.
Among the larger Rodinia kernels, \textit{lavamd} reaches ${\sim}8\times$
speedup at moderate area.

\paragraph{Intermediate and non-monotonic behavior}
For PRESENT, NW, and DES, scaling improves latency overall but the
progression is not strictly monotonic: in several cases
(e.g., PRESENT, DES), $N{=}8$ yields Pareto-optimal designs not dominated
by $N{=}10$, reflecting stochastic exploration effects.
\textit{Leukocyte} presents a distinct pattern: its top speedup
(${\sim}14$--$15\times$) originates from a baseline optimization that
agents do not replicate, though they do improve designs in the
${\sim}7$--$10\times$ range.

\paragraph{Saturation under simplicity or tight resource budgets}
Simpler or resource-constrained kernels exhibit diminishing returns.
\textit{KMP} saturates early (${\sim}10$--$12\times$), and \textit{AES}
shows similar performance for $N{=}8$ and $N{=}10$.
\textit{NW}, which operates near its area limit (${\sim}10$K), illustrates
a related effect: $N{=}8$ slightly outperforms $N{=}10$, suggesting that
under tight area budgets additional agents may explore configurations
that trade area for diminishing latency gains.

\paragraph{Global optimization beyond sub-kernel decomposition}
Across benchmarks, the best final designs do not always originate from the
top-ranked ILP variant. This confirms that Stage~2 optimization---through
pragma recombination and code-level transformations applied across function
boundaries---can exploit cross-function interactions that the ILP model,
which estimates global latency from isolated sub-kernel synthesis, cannot
capture.

\begin{figure}
    \centering
    \includegraphics[width=0.8\linewidth]{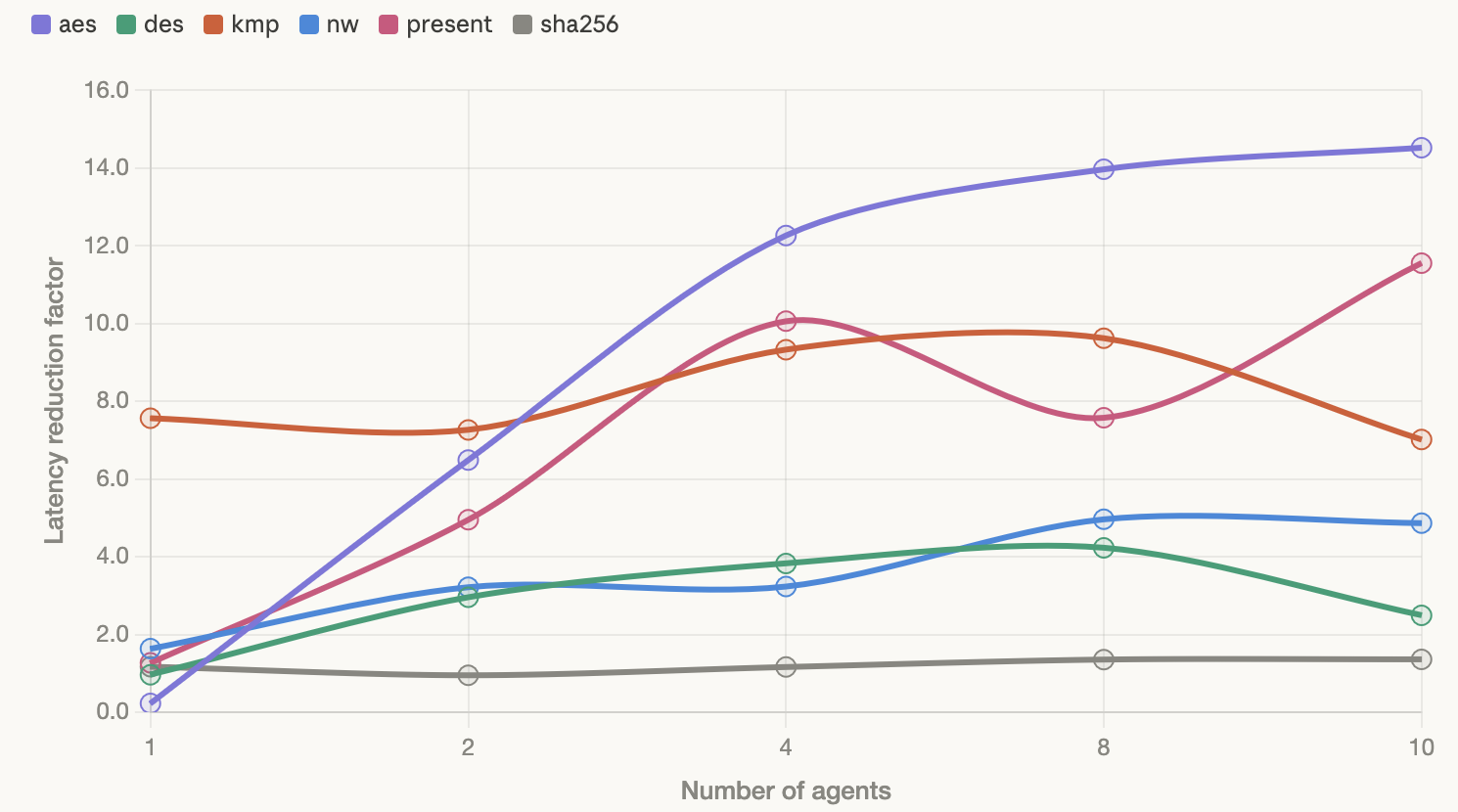}
    \caption{Latency improvement factor over baseline versus number of expert agents across six benchmarks. Improvements range from 1.4$\times$ to 14.5$\times$ and generally increase with agent count, typically plateauing after four agents.}
    \label{fig:abc-latency}
    \vspace{-10pt}
\end{figure}

\subsection{Generalization to ASICs}
The approach presented in this paper uses HLS tools with FPGA as the target device 
to evaluate and guide the agent-based optimization flow. While the approach itself 
is tool-agnostic, and can use ASIC-targeted HLS and logic synthesis tools to evaluate 
and guide the flow, we show that both latency improvements and area correlate well to 
an ASIC-mapped design as well.

To measure how well area reported by HLS tool correlates with ASIC-mapped area, we compute correlation between HLS-reported synthesis area and logic area obtained from the open-source logic synthesis tool ABC, across the six HLS-Eval benchmarks. The strength of the correlation varies: SHA256 ($r{=}0.992$), KMP ($r{=}0.964$), and NW ($r{=}0.859$) show strong near-linear relationships, indicating that HLS area estimates are generally reliable proxies for silicon cost on these kernels. AES ($r{=}0.757$) shows moderate-to-strong correlation, while DES ($r{=}0.603$) is moderate. PRESENT ($r{=}0.277$) exhibits weak correlation, suggesting that HLS area estimates may be less reliable for this design. This variation is because HLS tool area estimates tend to be inaccurate for designs with more memory instances.

Figure~\ref{fig:abc-latency} shows the factor of improvement in latency (compared to the baseline design) on the y-axis and number of agents on the x-axis. Latency is computed as the product of number of cycles obtained from the HLS timing report and an estimate for clock period obtained from ABC synthesis report. The latency value for different number of agents is then normalized by the latency of the baseline design (without agentic optimization) to get the latency reduction factor on the y-axis. Across the six benchmarks, latency can be improved by a factor ranging from 1.4X to 14.5X. For all benchmarks, latency improvements increase with increasing number of expert agents, typically plateauing after four expert agents.

\subsection{Discussion and Future Work}

This work presents an initial exploration of agent scaling for HLS design space
exploration. Our results demonstrate that \textit{Agent Factories}, when applied to High Level Synthesis domain yield several  consistent trends: increasing the number of
agents improve exploration and extend the Pareto frontier on several 
kernels. This suggests that agent scaling, analogous to inference-time scaling in
LLMs can serve as a powerful mechanism for navigating large hardware designs
spaces without explicit training or handcrafted heuristics. While these results are encouraging, the current study can be significantly expanded in scope. Our evaluation spans twelve benchmarks and focuses on Vitis HLS with logic-level validation through ABC, without yet incorporating broader benchmark suites (e.g., HLSyn) or comparisons to advanced automated DSE frameworks such as AutoDSE. 

Several directions follow naturally from this observation: incorporating learning
(e.g., fine-tuning or reinforcement learning) to improve generalization and
efficiency; expanding evaluation across broader benchmarks and toolchains,
including downstream synthesis; and integrating advanced agentic strategies
(e.g., memory, replay, coordination) to improve search quality. In addition, we are working towards several baselines comparisons with traditional optimization methods as well for HLS design space exploration to determine the differentiation and tradeoffs involved with Agent Factoties.

This research demonstrates a powerful result that taken together, the direction point toward a broader vision in which
agent-based systems, combined with learning and deeper integration into hardware
toolchains, act as scalable optimizers for complex design spaces, reducing the
need for manual expertise and enabling more automated hardware design.

\appendices

\section{Agent Scaling Token Usage}

\begin{figure}[htbp]
\centering
\includegraphics[width=\linewidth]{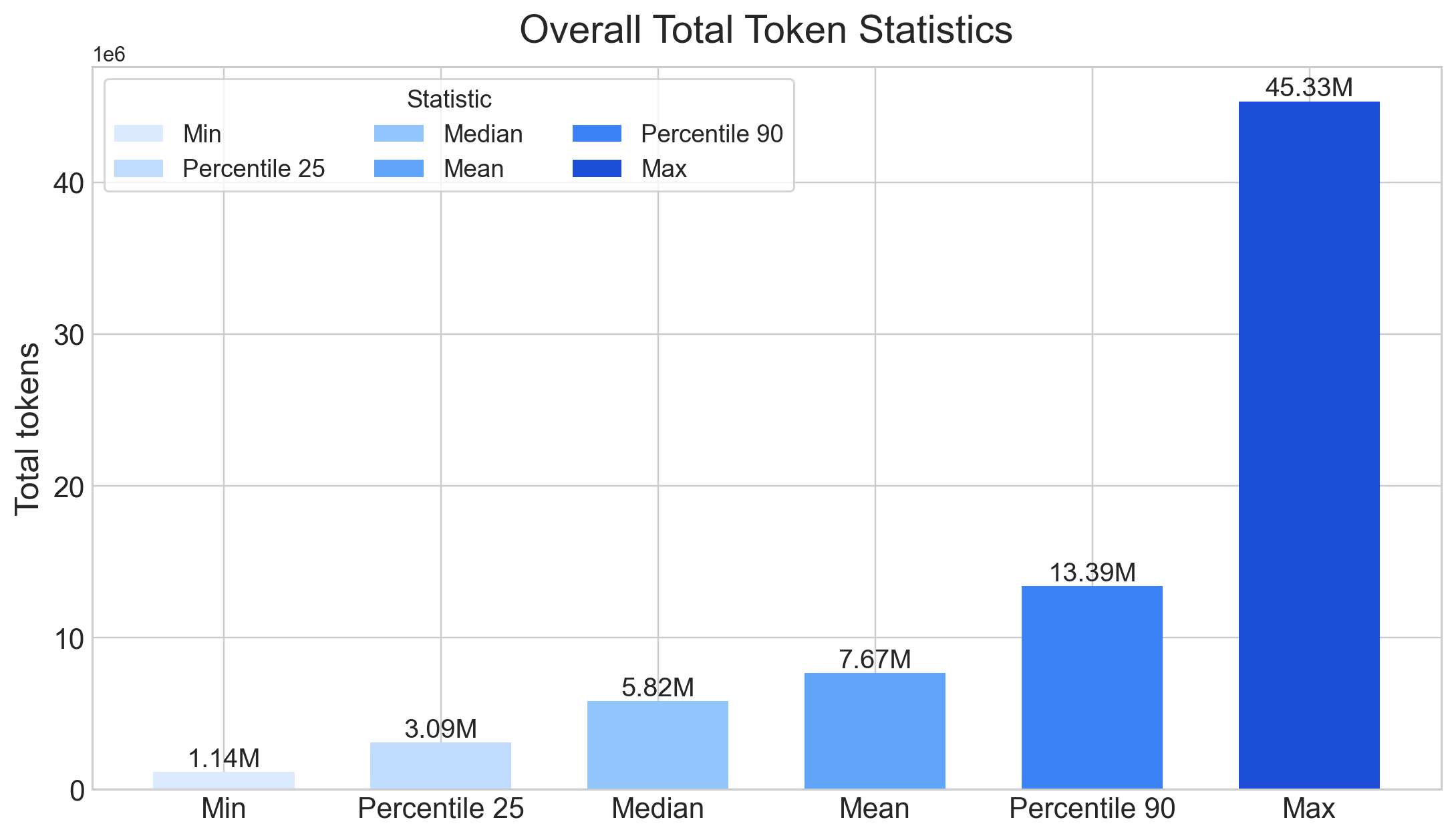}
\caption{Average Inference cost of agent scaling. Each session is combination of Opus 4.5/4.6.}
\label{fig:total-token-stats}
\vspace{-10pt}
\end{figure}

Token usage is reported as the total number of model tokens consumed by the
agents during execution, with Claude Opus~4.5 and Claude Opus~4.6 used as
the underlying models. Across all valid runs, the method consumed a median
of 5.82M~tokens and a mean of 7.67M~tokens per run, indicating a
right-skewed distribution with a small number of substantially more expensive
runs. The 25th and 90th percentiles were 3.09M and 13.39M~tokens,
respectively, while the observed range spanned from 1.14M to 45.33M~tokens.
These results indicate that token consumption is typically in the low-to-mid
millions per run, with occasional substantially higher-cost outliers. The
overall token summary is shown in Fig.~\ref{fig:total-token-stats}.

\section{Algorithm}
Algorithm~\ref{alg:pipeline} summarizes the complete two-stage pipeline.
By parameterizing Stage~2 with $N$, we study how increasing the number of
parallel exploration agents affects solution quality.
Larger $N$ expands the explored region of the design space, increasing the
probability of discovering lower-latency implementations.
We evaluate $N\in\{1,2,4,8,10\}$ in Section~\ref{sec:results}.

\begin{algorithm}[t]
\caption{Two-Stage Multi-Agent Design Space Exploration for HLS}
\label{alg:pipeline}
\KwIn{Design $\mathcal{D}$, area budget $\mathcal{A}_{\max}$, agents $N$}
\KwOut{Optimized design $\mathcal{D}^*$}

Coordinator extracts call graph $G=(\mathcal{F},\mathcal{E})$ from $\mathcal{D}$\;

\ForAll{$f_k \in \mathcal{F}$ \textbf{in parallel}}{
    launch optimizer agent $a_k$\;
    $\mathcal{V}_k \gets a_k.\textsc{SearchAndEvaluate}(f_k)$\;
}

ILP agent builds a global model from $\{\mathcal{V}_k\}$ and $G$\;
$\mathcal{S} \gets$ top-$N$ feasible ILP solutions\;

\ForAll{$s_i \in \mathcal{S}$ \textbf{in parallel}}{
    launch exploration agent $g_i$\;
    $\mathcal{R}_i \gets g_i.\textsc{TargetedRefinement}(\mathcal{D}, s_i, \mathcal{A}_{\max})$\;
}

$\mathcal{D}^* \gets \arg\min_{d \in \cup_i \mathcal{R}_i} L(d)
\;\text{s.t.}\; A(d)\leq \mathcal{A}_{\max}$\;
\Return{$\mathcal{D}^*$}\;

\SetKwProg{Fn}{Function}{:}{}

\Fn{\textsc{SearchAndEvaluate}($f_k$)}{
    generate candidate variants $\{v_k^m\}_{m=0}^{M}$\;
    $\mathcal{V}_k \gets \{(v_k^m,L^m,A^m)\mid \textsc{Correctness}(v_k^m)\}$\;
    \Return{verified variants in $\mathcal{V}_k$ with HLS metrics}\;
}

\Fn{\textsc{TargetedRefinement}($\mathcal{D}, s, \mathcal{A}_{\max}$)}{
    instantiate $\mathcal{D}_s \gets \textsc{Instantiate}(\mathcal{D},s)$\;
    generate refinement attempts $\{p_j\}$\;
    $\mathcal{R} \gets \{(\mathcal{D}_j,L^j,A^j)\mid \mathcal{D}_j=\textsc{Apply}(p_j,\mathcal{D}_s),\,
    \textsc{Correctness}(\mathcal{D}_j),\, A^j\leq \mathcal{A}_{\max}\}$\;
    \Return{$\mathcal{R}$ if non-empty, else $\{(\mathcal{D}_s,L_s,A_s)\}$}\;
}
\end{algorithm}

\bibliographystyle{IEEEtran}
\bibliography{references}

\end{document}